\PassOptionsToPackage{unicode}{hyperref}
\PassOptionsToPackage{hyphens}{url}
\documentclass[
  a4paper,
]{article}
\usepackage{amsmath,amssymb}
\usepackage{iftex}
\ifPDFTeX
  \usepackage[T1]{fontenc}
  \usepackage[utf8]{inputenc}
  \usepackage{textcomp} 
\else 
  \usepackage{unicode-math} 
  \defaultfontfeatures{Scale=MatchLowercase}
  \defaultfontfeatures[\rmfamily]{Ligatures=TeX,Scale=1}
\fi
\usepackage{lmodern}
\ifPDFTeX\else
\fi
\IfFileExists{upquote.sty}{\usepackage{upquote}}{}
\IfFileExists{microtype.sty}{
  \usepackage[]{microtype}
  \UseMicrotypeSet[protrusion]{basicmath} 
}{}
\makeatletter
\@ifundefined{KOMAClassName}{
  \IfFileExists{parskip.sty}{%
    \usepackage{parskip}
  }{
    \setlength{\parindent}{0pt}
    \setlength{\parskip}{6pt plus 2pt minus 1pt}}
}{
  \KOMAoptions{parskip=half}}
\makeatother
\usepackage{xcolor}
\usepackage{color}
\usepackage{fancyvrb}

\DefineVerbatimEnvironment{Highlighting}{Verbatim}{commandchars=\\\{\}}
\newenvironment{Shaded}{}{}

\newcommand{\NormalTok}[1]{#1}

\usepackage{longtable,booktabs,array}
\usepackage{calc} 
\usepackage{etoolbox}
\makeatletter
\patchcmd\longtable{\par}{\if@noskipsec\mbox{}\fi\par}{}{}
\makeatother
\IfFileExists{footnotehyper.sty}{\usepackage{footnotehyper}}{\usepackage{footnote}}
\makesavenoteenv{longtable}
\usepackage{graphicx}
\makeatletter
\def\maxwidth{\ifdim\Gin@nat@width>\linewidth\linewidth\else\Gin@nat@width\fi}
\def\maxheight{\ifdim\Gin@nat@height>\textheight\textheight\else\Gin@nat@height\fi}
\makeatother
\setkeys{Gin}{width=\maxwidth,height=\maxheight,keepaspectratio}
\makeatletter
\def\fps@figure{htbp}
\makeatother
\providecommand{\tightlist}{%
  \setlength{\itemsep}{0pt}\setlength{\parskip}{0pt}}
\NewDocumentCommand\citeproctext{}{}

\makeatletter
 \let\@cite@ofmt\@firstofone
 \def\@biblabel#1{}
 \def\@cite#1#2{{#1\if@tempswa , #2\fi}}
\makeatother
\newlength{\cslhangindent}
\newlength{\csllabelwidth}
\newenvironment{CSLReferences}[2] 
 {\begin{list}{}{%
  \setlength{\itemindent}{0pt}
  \setlength{\leftmargin}{0pt}
  \setlength{\parsep}{0pt}
  \ifodd #1
   \setlength{\leftmargin}{\cslhangindent}
   \setlength{\itemindent}{-1\cslhangindent}
  \fi
  \setlength{\itemsep}{#2\baselineskip}}}
 {\end{list}}
\usepackage{calc}

\ifLuaTeX
\usepackage[bidi=basic]{babel}
\else
\usepackage[bidi=default]{babel}
\fi
\babelprovide[main,import]{american}

\def\languageshorthands#1{}
\makeatletter
\@ifpackageloaded{subfig}{}{\usepackage{subfig}}
\@ifpackageloaded{caption}{}{\usepackage{caption}}
\AtBeginDocument{%

}
\AtBeginDocument{%

}
\newcounter{pandoccrossref@subfigures@footnote@counter}
{\end{figure}%
\addtocounter{footnote}{-\value{pandoccrossref@subfigures@footnote@counter}}
\@for\f:=\global@pandoccrossref@subfigures@footnotes\do{\stepcounter{footnote}\footnotetext{\f}}%
\gdef\global@pandoccrossref@subfigures@footnotes{}}
\@ifpackageloaded{float}{}{\usepackage{float}}
\floatstyle{ruled}
\@ifundefined{c@chapter}{\newfloat{codelisting}{h}{lop}}{\newfloat{codelisting}{h}{lop}[chapter]}
\floatname{codelisting}{Listing}

\makeatother
\ifLuaTeX
  \usepackage{selnolig}  
\fi
\usepackage{bookmark}
\IfFileExists{xurl.sty}{\usepackage{xurl}}{} 
\hypersetup{
  pdftitle={NER4all or Context is All You Need},
  pdfauthor={Torsten Hiltmann1,; Martin Dröge1,3; Nicole Dresselhaus1,2; Till Grallert1,2,; Melanie Althage1; Paul Bayer1; Sophie Eckenstaler1,4; Koray Mendi1; Jascha Marijn Schmitz1,2; Philipp Schneider1; Wiebke Sczeponik1; Anica Skibba1,2},
  pdflang={en-US},
  pdfkeywords={large language models, named entity
recognition, history},
  hidelinks,
  pdfcreator={LaTeX via pandoc}}

\title{NER4all or Context is All You Need}
\usepackage{etoolbox}
\makeatletter
\providecommand{\subtitle}[1]{
  \apptocmd{\@title}{\par {\large #1 \par}}{}{}
}
\makeatother
\subtitle{Using LLMs for low-effort, high-performance NER on historical
texts. A humanities informed approach.}
\author{Torsten Hiltmann\textsuperscript{1,*} \and Martin
Dröge\textsuperscript{1,3} \and Nicole
Dresselhaus\textsuperscript{1,2} \and Till
Grallert\textsuperscript{1,2,*} \and Melanie
Althage\textsuperscript{1} \and Paul
Bayer\textsuperscript{1} \and Sophie
Eckenstaler\textsuperscript{1,4} \and Koray
Mendi\textsuperscript{1} \and Jascha Marijn
Schmitz\textsuperscript{1,2} \and Philipp
Schneider\textsuperscript{1} \and Wiebke
Sczeponik\textsuperscript{1} \and Anica Skibba\textsuperscript{1,2}}
\date{30 January 2025}

\begin{document}
\maketitle
\begin{abstract}
Named entity recognition (NER) is a core task for historical research in
automatically establishing all references to people, places, events and
the like. Yet, do to the high linguistic and genre diversity of sources,
only limited canonisation of spellings, the level of required historical
domain knowledge, and the scarcity of annotated training data,
established approaches to natural language processing (NLP) have been
both extremely expensive and yielded only unsatisfactory results in
terms of recall and precision. Our paper introduces a new approach. We
demonstrate how readily-available, state-of-the-art LLMs significantly
outperform two leading NLP frameworks, spaCy and flair, for NER in
historical documents by seven to twentytwo percent higher F1-Scores. Our
ablation study shows how providing historical context to the task and a
bit of persona modelling that turns focus away from a purely linguistic
approach are core to a successful prompting strategy. We also
demonstrate that, contrary to our expectations, providing increasing
numbers of examples in few-shot approaches does not improve recall or
precision below a threshold of 16-shot. In consequence, our approach
democratises access to NER for all historians by removing the barrier of
scripting languages and computational skills required for established
NLP tools and instead leveraging natural language prompts and
consumer-grade tools and frontends.
\end{abstract}

\textsuperscript{1} Humboldt-Universität zu Berlin\\
\textsuperscript{2} NFDI4Memory\\
\textsuperscript{3} AI-Skills\\
\textsuperscript{4} Kompetenzwerkstatt Digital Humanities

\textsuperscript{*} Correspondence:
\href{mailto:torsten.hiltmann@hu-berlin.de}{Torsten Hiltmann
\textless{}torsten.hiltmann@hu-berlin.de\textgreater{}},
\href{mailto:till.grallert@hu-berlin.de}{Till Grallert
\textless{}till.grallert@hu-berlin.de\textgreater{}}

\section{Plain language summary}\label{plain-language-summary}

This paper introduces and positively evaluates a new method for
{\emph{named entity recognition}} ({NER}), or identifying and
classifying references to real-world entities, such as people or places,
in historical texts. {NER} is a foundational first task for many
historical research research questions as it allows us to screen large
bodies of textual sources for relevant entities of interest. Yet, source
corpora for historical research are commonly highly diverse in language,
genre, and structure and very different from the modern texts with their
highly regular language and standardised orthography. Thus, common {NER}
tools, trained on modern texts, perform rather badly for our use case
scenario without extensive and expensive pre-processing, customisation,
and retraining of models.

In this paper we show how out-of-the-box, commercial large language
models ({LLM}s) significantly outperform established frameworks for
{NER} using natural language prompts in both German and English. We
argue that in order to so, one has to reconceptualize {NER} from a
purely linguistic task into a humanist endeavour that requires some
level of domain expertise and aims at activating the vast body of
information {LLM}s have ingested during their training.

To test our approach, we created ground truth with manually annotated
named entities from the 1921 Baedeker travel guide for Berlin and
surroundings and evaluated the impact of various strategies for
prompting readily available, commercial {LLM}s, namely ChatGPT-4o, on
identifying ({\emph{recall}}) and classifying ({\emph{precision}}) named
entities. Prompting strategies comprise the provision of contextual
information and persona modelling directing the {LLM} away from a purely
linguistic approach to {NER}, providing increasing numbers of random
examples from our corpus (zero, few, and many-shot), and common prompt
engineering techniques, such as reiterating instructions, offering
rewards or punishments, and insisting on slow and thorough thought
processes. Finally we compared all results to baselines generated with
two leading {NER} frameworks, flair and spaCy.

Our results demonstrate that {LLM}s consequently perform at least on par
with flair and spaCy and significantly outperform such traditional {NER}
frameworks as soon as a bit of contextual information and persona
modelling is included in the prompts. We also show that, surprisingly
and against our expectations, zero-shot approaches, that is prompts
without any examples, perform better than few-shot approaches until the
number of examples reaches 16 and more. Examples require annotation from
domain experts and are thus expensive. Including them in the prompt also
increases the necessary context-window, which requires more computing
power and has to be paid for for when using commercial {LLM}s, as we
did. Finally, we could show that although initially English performs
better than German as a prompting language in terms of {\emph{recall}},
this difference becomes insignificant when we incorporated our rather
simple prompt engineering techniques.

\section{Conflict of interest
statement}\label{conflict-of-interest-statement}

Competing interests: The author(s) declare none

\section{Introduction}\label{introduction}

The digital has brought in a new era of abundance for historical
research and with it new challenges in order to make sense of vast
amounts of highly diverse and unstructured data encapsulated in
digitised---and increasingly born-digital---cultural artifacts (c.f.
Rosenzweig 2003). Named entity recognition ({NER}), or the ability to
automatically identify objects of interest for the historian, is one of
the fundamental computational approaches for information extraction.
Yet, our approaches to {NER} have hitherto been adopted from the domains
of natural language processing ({NLP}) and computational linguistics.
Existing tools work best with homogenous corpora of normalized texts in
modern languages and come with steep learning curves for humanists
without prior computing knowledge. Adapting and applying them to the
specific challenges of historical corpora, requires prohibitive
expenditures in effort, time, and money.

The recent hype around Large Language Models ({LLM}) has led to the
expectation that they will solve historical {NER} with aplomb, but so
far the performance of {LLM} is still significantly lower than
established machine-learning approaches, such as transformer
architectures (Keraghel, Morbieu, and Nadif 2024). In this paper we show
demonstrate how rethinking prompting strategies and fundamentally
reconceptualising {NER} from a linguistic task to a humanities-focused
task with contextual information and appropriate persona modelling,
allows us to significantly outperform state-of-the-art {NLP} tools such
as spaCy and even flair in recall and precision using common
foundation-models and 0-shot prompts without further training or
fine-tuning and without creating additional data. We thus fundamentally
remove barriers and democratize access to highly performant {NER} on
heterogeneous corpora of very specific historical and frequently
low-resourced texts.

\subsection{The challenge of NER for historical
texts}\label{the-challenge-of-ner-for-historical-texts}

Named entities are references to distinct and uniquely nameable or
designatable concepts that can be either concrete or abstract (Ehrmann
2008). Named Entity Recognition ({NER}) is a Natural Language Processing
({NLP}) task to detect these references in unstructured text data and
classify them according to predefined categories, such as people, places
or organizations. {NER} is thus a token classification task with a
desired result like the following: {\emph{t\textsubscript{1},
t\textsubscript{2}: PER (person), t\textsubscript{3}: PER (person),
t\textsubscript{4} \ldots{} t\textsubscript{9}, t\textsubscript{10}: LOC
(location), t\textsubscript{11}: LOC (location), t\textsubscript{12},
\ldots{}}} However, the disambiguation or identification of these named
entities is not part of the {NER} task itself, but represents two
subsequent processing steps({NE} disambiguation and linking).

While some domains operate with relatively homogeneous text data, such
as scientific publications, patient reports, or annual economic reports,
which are written in modern standardized language and often follow a
common internal structure, resulting in effective {NER} performance,
this is not the case for the historical humanities. Historical research
is characterized by texts defined by their diversity in form and
content, presenting a significant challenge for {NLP}-tasks. This
diversity may include:

\begin{itemize}
\tightlist
\item
  Language

  \begin{itemize}
  \tightlist
  \item
    Multilinguality of the documents we work with,
  \item
    Different levels of historical language, with documents dating from
    different periods of time (e.g., Middle English, Early Modern
    English, Classical Latin, Medieval Latin) and varying dialects,
  \item
    Lack of standardized spelling in older texts and changing
    orthographic conventions that reflect linguistic and cultural shifts
    over centuries,
  \item
    Different editorial treatments of the texts, such as diplomatic and
    critical editions or normalised reading versions (e.g.~providing the
    source text as it appears in the sources or reworking it according
    to certain rules).
  \end{itemize}
\item
  Text structure and vocabulary

  \begin{itemize}
  \tightlist
  \item
    Different text genres (chronicles, charters, diary entries, letters,
    press articles, \ldots) ,
  \item
    Changing cultural customs, such as forms of personal address, use of
    titles, and other social conventions,
  \item
    Different domains (law, economy, religion, military, food culture,
    etc.),
  \item
    Changes over time in the vocabulary and meanings of words.
  \end{itemize}
\item
  Data

  \begin{itemize}
  \tightlist
  \item
    Scarcity of resources and limited annotated data.
  \end{itemize}
\end{itemize}

Moreover, the process of detecting and classifying named entities in
historical texts is often less straightforward than it appears as it
involves a degree of interpretation by domain experts familiar with the
specific subject, historical context, and time period. In their reliance
on semi-automated workflows, historians, therefore, {\textbf{value
recall over precision}}.

{NER} is a critical first step in accessing and analyzing unknown texts
for historical scholarship. Identifying that and which persons, places,
or institutions are mentioned in a given text is essential for
historians seeking to understand and interpret new sources and to find
new information about different entities that may be mentioned in the
text. Consequently, published editions and other tools for accessing
historical sources have always included indexes of persons, places, and
subjects as a central and integral part.

Automated {NER} has become even more important in the face of major
digitization campaigns that have made millions of textual documents from
archives and libraries available as scans (e.g.,
\href{https://www.archivportal-d.de/}{Archivportal-D}) and are
increasingly being converted to full text via Optical Character
Recognition ({OCR}) or Handwritten Text Recognition ({HTR}). However,
the challenges our sources pose to {NER}, as outlined above, and from
commonly trained on contemporary language and genres (particularly news
texts, social media, and internet fora), have made automated {NER} with
consistently high-quality results in both recall and precision nigh
impossible

Historians, therefore, have to adopt one of the following strategies if
they want to apply {NER}. All three involve a significant degree of
compromise or a considerable investment of time and resources.

\begin{enumerate}
\def\labelenumi{\arabic{enumi}.}
\tightlist
\item
  Use of ready-made models for general tag sets, resulting in a
  patchwork of results for historical texts. Performance correlates with
  similarity of historical input to training corpora.
\item
  Adapting the input texts to the models to address this under
  performance through normalization, i.e.~standardized, modern spelling.
  Problems caused by different genres and contexts remain (Ehrmann et
  al. 2023).
\item
  Generating a tailored rule set or acquiring sufficient training data
  to refine existing models or train an entirely new one. This yields
  the highest quality results, but is also the most resource-intensive
  and out of reach for many applications. Its efficacy is contingent
  upon the availability of sufficient data for annotation to justify the
  investment of resources Y. Chen et al. (2015) and case-specific domain
  expertise. In some cases, automated processing may even not be
  feasible at all due to the limited number of texts and data available.
\end{enumerate}

With this paper, we demonstrate that these problems can be solved
through the appropriate use of Large Language Models. Using our
approach, {LLM}s can provide agnostic {NER} annotations for any type of
historical text and potentially any tag set without the need to create
specific training data and train or fine-tune a specific model. This
represents a significant advancement in the automated analysis of
historical documents, thereby transforming the field.

\subsection{Current state of the
field}\label{current-state-of-the-field}

{NER} techniques roughly fall into two camps (Keraghel, Morbieu, and
Nadif 2024; Pakhale 2023): rule-based heuristics and machine-learning
approaches. Deep-Learning approaches to {NER}, in particular in the form
of fine-tuned variations of transformer-based language models and
especially the {BERT}-model, have in recent years become the most widely
discussed approach to {NER} on historical text, due to their high
performance if purposefully and extensively trained on specific sources
(Ehrmann et al. 2023, 23ff).

This has been radically changed with the publication of ChatGPT in
November 2022, as we will argue in this paper. {LLM}s have rapidly
gained popularity in historical research as prospective multipurpose
tools that, due to their much improved complexity and domain
indifference, might not just alter the tradeoffs but remove them
altogether. Applications range from {OCR}/{OCR} correction (e.g. Thomas,
Gaizauskas, and Lu 2024; van Eijnatten 2024), to (semi-)automatic
building of knowledge graphs (e.g. Giovanelli and Traviglia 2024;
Graham, Yates, and El-Roby 2023), and {NLP} tasks (e.g. Karjus 2023;
Stammbach, Antoniak, and Ash 2022), including {NER}. Approaches range
from zero-shot (De Toni et al. 2022; González-Gallardo et al. 2023;
Sarker et al. 2024), to single-shot (Santos et al. 2024; Tang et al.
2024), and few-shot prompting (Sarker et al. 2024; Tang et al. 2024)
using a wide array of proprietary and open {LLM}s: ChatGPT-4o (OpenAI
2024), GPT-3.5 / ChatGPT 3.5 (OpenAI 2022), Llama 2 (Touvron, Martin,
and Stone 2023), Alpaca 7B (Taori et al. 2023), t0 (Sanh et al. 2022),
and ChatGLM2-6B (THUDM 2023). Interestingly, these studies use {LLM}s
for heterogeneous linguistic domains the models were not originally
trained on, demonstrating the specific challenges for {NER} in the field
and the transfer learning capabilities of {LLM}s, which makes them
promising candidates for historical {NER}:
18\textsuperscript{th}--19\textsuperscript{th} century newspapers
(English, French, German) (De Toni et al. 2022);
19\textsuperscript{th}--20\textsuperscript{th} century newspapers and
commentaries on classical greek literature (German, French,
Luxembourgish, English, Finnish, Swedish) (González-Gallardo et al.
2023); 18\textsuperscript{th} century parish reports (Portuguese)
(Santos et al. 2024); 17\textsuperscript{th} century notary records
(Spanish) (Sarker et al. 2024); 18\textsuperscript{th} century
historical chronicles (Classical Chinese) (Tang et al. 2024).

Due to rapidly evolving technologies and the frequent publication of new
and ever larger models, such applications are still in an early and
experimental phase. Most studies evaluated results by comparing their
results with established {NER}-tools (e.g.,the embeddings and
transformer-based {NLP}-framework flair (Akbik et al. 2019) or various
transformer based Language Models, such as fine-tuned {BERT} models). In
all cases where comparative tests were conducted, {LLM}s performed worse
in many or even all tasks they were given, especially compared to
{BERT}-models. Based on those largely negative results,
(González-Gallardo et al. 2023; Sarker et al. 2024; Tang et al. 2024)
came to unfavorable conclusions regarding LLMs capabilities for {NER} on
historical texts, although the latter observed good performances for
other NLP tasks (particularly Relationship Extraction). The other
studies remained cautious until further and more extensive
experimentation (De Toni et al. 2022) or concluded to see LLMs
potentials in a more limited, assisting role (Santos et al. 2024).

Crucially, Tang et al. (2024) are the only ones explaining their usage
of advanced techniques of prompt engineering targeting {LLM}s in-context
learning capabilities, achieving good results for Relationship
Extraction based on {NER} with some {LLM}s. Similarly Holla, Kumar, and
Singh (2024) highlight the role of context to further improve
benchmarks, although from the angle of adding additional tasks to the
training procedure. Evaluating prompting strategies, including the
structure of prompts and attempts at prompt engineering in a systematic
way is therefore difficult, as the published documentation is usually
insufficient or altogether lacking.

In most instances, {LLM}s have been used out-of-the-box, assuming they
``understand'' the prompt and can do the job. Opponents of such
approaches argue that {LLM}s are purely predicting the most likely next
token based on the input and the vast amount of training data and are
therefore of limited use for sequence annotation. Wang et al. (2023)
tried to solve this problem by introducing special sequences into the
text to annotate named entities. In addition, they improved the
performance of their {LLM} for {NER} by optimizing the structure and
information content of the examples for few-shot learning.

Against this backdrop, we argue that {LLM}s have not yet been used to
their full potential for common tasks in historical research, based,
mainly, on insights from the wider {LLM}-research community regarding
advanced prompt engineering, such as chain-of-thought prompting, and
{LLM}s' in-context learning capabilities, in particular of newer
multimodal models that are not limited to texts, such as GPT-4o (e.g.,
Zixiao et al. 2024; Wang et al. 2023; Brown et al. 2020; J. Chen et al.
2023; Levy, Bogin, and Berant 2023).

\section{Shifting the perspective from linguistics to history and
contextual
information}\label{shifting-the-perspective-from-linguistics-to-history-and-contextual-information}

We propose a paradigmatic shift in the use of {LLM}s for {NLP} tasks:
the redefinition of these tasks from a purely linguistic dimension to a
content-oriented humanities dimension. Instead of treating the task as a
technical exercise in {NLP}, we argue for an approach that emphasizes
the contextual and interpretative needs for solving corresponding tasks
in the humanities. This means that we should use {LLM}s not only to
process and recognize linguistic structures but also to analyze their
content during {NER} tasks. As we show below, we can overcome the unique
challenges presented by historical documents and other complex texts in
the humanities and gain more accurate and meaningful insights following
such an approach.

Previous approaches have focused primarily on the emergent abilities
that result from the scaling process of {LLM}s. These skills include the
execution of specific instructions without additional training, known as
in-context learning, including instruction following, step-by-step
reasoning, and so on. However, scaling language models to large language
models by training them on larger amounts of texts entails not only
these abilities, but also the inclusion of a large amount of information
represented in the training texts.

We argue that through this process, the models not only learned new
tasks, but also learned the contextual information necessary to perform
these tasks and make more informed decisions. This means that {LLM}s
possess both the ability to perform the task on a linguistic level as
well as the necessary contextual information {NER} tasks.

While effective prompting is crucial to mobilize and exploit these new
capabilities, we believe that the integration of contextual information
embedded in the model during training is the key determinant of the
success of {LLM}s for tasks such as {NER}. This embedded knowledge
allows the model to perform tasks with appropriate contextual
understanding, ultimately mimicking human behaviour in annotation
processes.\footnote{By using anthropomorphic language such as
  ``understanding'' and ``knowledge'', we do not aim at insinuating
  sentience.} As mentioned above, human annotators equally require
extensive domain knowledge to perform these tasks correctly and to make
informed decisions. They base their decisions not only on linguistic
cues, but also on their extensive knowledge of the subject matter.

The presence of this contextual knowledge in {LLM}s is evidenced by
studies such as OpenAI's GPT-4 technical paper, according to which GPT-4
achieves over 80\% accuracy on unspecified factual questions in history,
demonstrating that the model has access to a broad base of historical
information present in its training data (OpenAI et al. 2024). Our
approach aims at making this embedded information accessible for
decision making by including relevant context in the prompt, thereby
activating the model's contextual knowledge.

Aligning the annotated output with the input data in order to prevent
``hallucinations'' is a computational problem that we address below.
However, the correctness of responses now depends not only on linguistic
NLP tasks but also on the contextual information we incorporate into the
prompt. By prominently incorporating this context into the prompt, we
ensure that it is integrated into the decision process.

\section{Corpus and data set}\label{corpus-and-data-set}

Since our goal was to test this {LLM}-based approach to {NER} under
real-world conditions and in actual application scenarios and as the
approach required some coherent historical corpus in order to provide
clearly specified contextual information, we settled on a text from our
own historical domain expertise instead of the usual benchmark texts. We
chose the 19\textsuperscript{th} edition of the Baedeker travel guide
for Berlin and surroundings (Baedeker and Graupe 1921) because it has a
well-defined historical context (information about traveling and
sightseeing in Berlin around 1920), partly uses period-specific
language, and also includes an interesting variety of structural
features from narrative sections to comprehensive lists, such as stops
on different bus routes. Travel guides are a good genre to benchmark
{NER} performance because of the density and variety of named entities
mentioned therein

The book was scanned and {OCR}'d with
\href{https://www.wikidata.org/wiki/Q124347709}{OCR4all} (Reul et al.
2019), which acts as a service wrapping various tools for layout and
text recognition into highly customizable workflows. We trained our own
text recognition model based on the ``idiotikon'' Calamari model from
the \href{https://www.idiotikon.ch/}{Schweizerisches Idiotikon} (Reul
and Wick 2021; Wick, Reul, and Puppe 2020) for 29 randomly selected
pages (12.55\% of the entire corpus), which resulted in a character
error rate ({CER}) of 0.0064.

Preprocessing for {NER} was limited to removing line breaks and
hyphenation and to normalizing of whitespace. Texts were then tokenized
with \href{https://www.wikidata.org/wiki/Q28406945}{spaCy} (Honnibal et
al. 2020).

The ground truth for {NER} was produced by manually annotating 55
randomly selected pages from our corpus with named entities using
{\href{https://www.wikidata.org/wiki/Q105824503}{INCEpTION}} (Klie et
al. 2018) following a set of guidelines specifically developed for this
corpus and based on the {NER} guidelines from the
\href{https://impresso.github.io/}{Impresso} project (Ehrmann et al.
2020). Each text was independently annotated by two team members.
Differences between annotations were reconciled based on a third
opinion.

We limited the types of entities to person (\texttt{PER}), organisation
(\texttt{ORG}), and places (\texttt{LOC}). Nested annotations were not
allowed. \texttt{PER} includes living people, historical and mythical
figures, as well as collectives dependent on their individual members,
such as the art collective ``Die Brücke''. Role names, salutations, and
addresses, such as ``geh{[}eimer{]} Regierungsrat'', were not included
in the \texttt{PER} tag. \texttt{ORG} differs from the kind of
collectives encoded as \texttt{PER} in being limited to incorporated
collectives, such as companies, public institutions, associations, etc.
\texttt{LOC} designates geographic locations. Given that most
incorporated collectives do have a physical location, consider, for
instance, a restaurant or even a city, our annotation as either
\texttt{ORG} or \texttt{LOC} depends on the context. If the text
foregrounds the locality, such as in the case of directions, we opted
for \texttt{LOC}. All other instances were annotated as \texttt{ORG}.
Out of this dataset we used 25 random pages for the evaluation (in
prodigy-format).

\section{Method}\label{method}

In this section, we describe our methodology for leveraging {LLM}s to
generate, evaluate, and refine named-entity annotations. We focus on the
technique of prompt engineering (Bsharat, Myrzakhan, and Shen 2024) to
achieve better results instead of optimizing training-datasets and
fine-tuning or retraining established approaches. This is due to the
prohibitive cost of the latter approaches and the performance of the
prompt-engineering approach as demonstrated below.

Because there are a lot of combinations for different prompting
techniques, we chose the best combination that worked during
experimentation without a comprehensive study. We then performed an
ablation study on this prompt to measure how much each component impacts
the overall performance---i.e.~what is the relative importance of the
prompt's parts for the observable output.

This is a very recent, radical changing field. While additionally having
slightly different effects on different {LLM}s, our experiments and
approach can only be based on the current state in an evolving
environment. More than any exact detail, we want to emphasise our main
thesis that having domain and input specific contextual information
plays an important role. We do not propose a definite set of
instructions on what to include and what not as gold standard.

Finally, we tested the impact of different prompting languages, namely
English and German.

In the following sections, we detail our chosen annotation format, which
adopts a span-based, tag-like structure inspired by prior research (e.g.
Wang et al. 2023). Next, we outline the instructions we provide to the
model---highlighting key challenges such as maintaining fidelity to the
original text, properly handling nested entities, and addressing
frequent labeling errors. We then discuss our matching strategy, which
employs fuzzy search techniques to robustly locate and compare generated
spans with their corresponding ground truth. To evaluate performance, we
measure precision, recall, and F\textsubscript{1}-Scores against two
established {NER} pipelines
(\href{https://www.wikidata.org/wiki/Q107386786}{flair} (Akbik et al.
2019) and \href{https://www.wikidata.org/wiki/Q28406945}{spaCy}
(Honnibal et al. 2020)), emphasizing prompt engineering over dataset
optimization or model retraining. We conclude by examining how one-shot,
few-shot, and many-shot examples affect annotation quality, and by
introducing the {LLM}s used in our experiments, notably
``gpt4o-2024-08-16'' (OpenAI 2024). Throughout, we stress the importance
of including domain-specific context and adapting prompts to an evolving
{LLM} landscape.

\subsection{Choice of target
annotations}\label{choice-of-target-annotations}

Similar to the approach of (Wang et al. 2023), we annotate complete
entity-spans, while also keeping in mind that it should be easily
reproducible by a generating {LLM}. Depending on the token-embedding
used we often saw generation of spaces at the end of a word, problems of
properly opening and closing of the tags, etc., even if the prompt
instructed the generation of a semantic tag for each word. As soon as we
switched over to tag-like annotation of spans of characters all these
problems nearly vanished even on smaller nets with smaller or
non-complete embeddings. Also we do not need to ``waste'' much of the
prompt on fixing the output-format. We suspect that this is due to a
high amount of similar structured text like code, html or math in the
training data.

We finally settled on
\texttt{\textless{}\textless{}TAG\ word\ word\ word\ /TAG\textgreater{}\textgreater{}}
as a way to annotate possibly nested spans and made all future
experiments with this format. Yan, Yu, and Chen (2024) reached similar
conclusions (having \texttt{\textless{}\textless{}},
\texttt{\textgreater{}\textgreater{}} perform best with a slightly
different layout).

\subsection{Instructing the LLM}\label{instructing-the-llm}

The final prompt (in both German and English) comprises multiple
components:

\begin{enumerate}
\def\labelenumi{\arabic{enumi}.}
\tightlist
\item
  A general {\textbf{introduction}} to the input text to be analyzed,
  with short, expert-level {\textbf{contextual information}}.
\item
  The main task and {\textbf{instructions}} on how to mark named
  entities in the text.
\item
  Some {\textbf{examples}} of input and expected output while making
  sure that all recognizable classes are included in decent amounts as
  Brown et al. (2020) demonstrated their significance for improving
  output quality.
\item
  A {\textbf{reiteration of the instructions}} in different wording and
  pointing out how in all examples these instructions were followed to
  the letter
\item
  Common tricks (Federiakin et al. 2024) to stress the importance of
  correctness and {\textbf{offer}} monetary {\textbf{reward}} for each
  successful classification
\item
  The seemingly important {\textbf{empty phrase}} ``take a deep breath
  and think step by step''
\end{enumerate}

\begin{figure}
\centering
\includegraphics{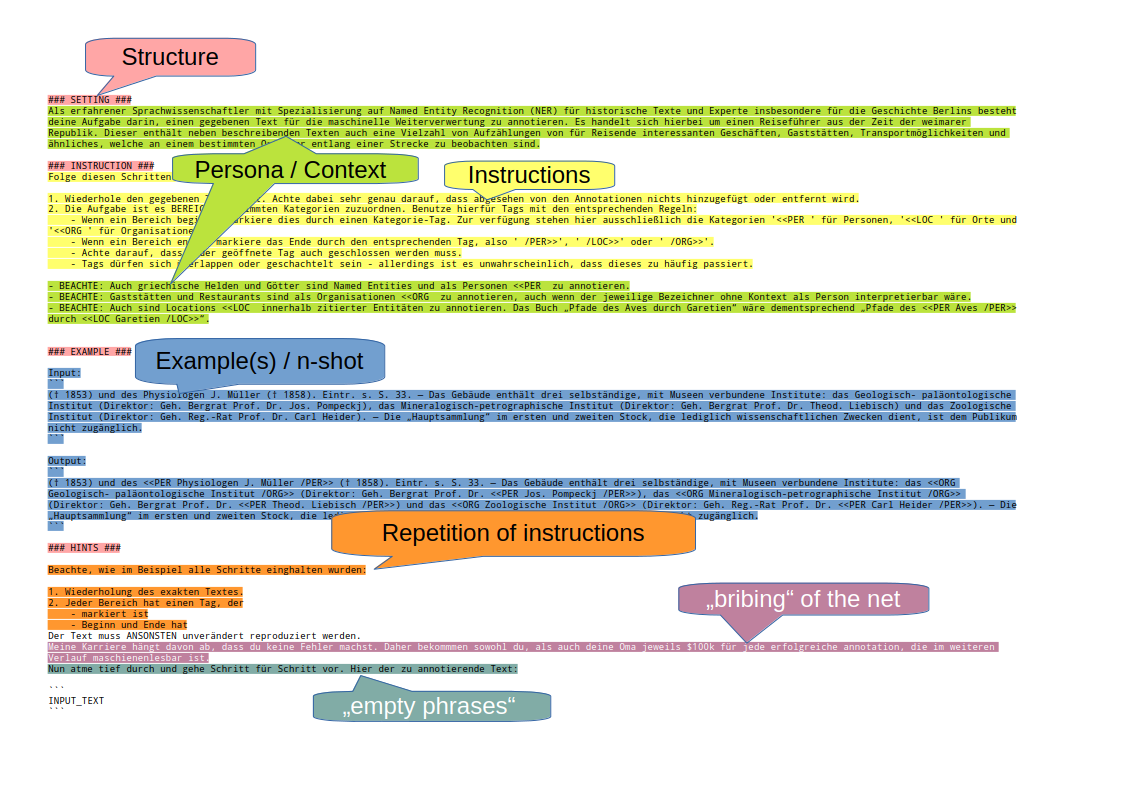}
\caption{Main instructions in the German prompt}\label{Figure 1}
\end{figure}

\subsection{Overview of the prompt
structure}\label{overview-of-the-prompt-structure}

The main instructions (2) are as follows (fig.1, note
that the screenshot shows the German prompt):

\phantomsection\label{prompt}%
\begin{Shaded}
\begin{Highlighting}[numbers=left,,]
\NormalTok{Follow these steps:}

\NormalTok{1.  Repeat the given text exactly. Be very careful to ensure that}
\NormalTok{    nothing is added or removed apart from the annotations.}
\NormalTok{2.  Your task is to categorize sections into predetermined categories.}
\NormalTok{    Use tags with corresponding rules:}
\NormalTok{    {-} When a section begins, mark it with a category tag. Available}
\NormalTok{      categories include \textquotesingle{}\textless{}\textless{}PER\textquotesingle{} for people, \textquotesingle{}\textless{}\textless{}LOC\textquotesingle{} for locations,}
\NormalTok{      and \textquotesingle{}\textless{}\textless{}ORG\textquotesingle{} for organizations.}
\NormalTok{    {-} When a section ends, mark the end with the corresponding tag,}
\NormalTok{      i.e., \textquotesingle{} /PER\textgreater{}\textgreater{}\textquotesingle{}, \textquotesingle{} /LOC\textgreater{}\textgreater{}\textquotesingle{}, or \textquotesingle{} /ORG\textgreater{}\textgreater{}\textquotesingle{}.}
\NormalTok{    {-} Be sure to close each opened tag.}
\NormalTok{    {-} Tags may overlap or be nested, but it is unlikely that this will}
\NormalTok{      happen frequently.}

\NormalTok{{-} NOTE: Even Greek heroes and gods are named entities and should be}
\NormalTok{  annotated as people (\textless{}\textless{}PER).}
\NormalTok{{-} NOTE: Restaurants are organizations (\textless{}\textless{}ORG) regardless of whether}
\NormalTok{  the name can be interpreted as a person without context.}
\NormalTok{{-} NOTE: Locations (\textless{}\textless{}LOC) within quoted entities should also be}
\NormalTok{  annotated. For example, "Pfade des Aves durch Garetien" would become}
\NormalTok{  "Pfade des \textless{}\textless{}PER Aves /PER\textgreater{}\textgreater{} durch \textless{}\textless{}LOC Garetien /LOC\textgreater{}\textgreater{}".}
\end{Highlighting}
\end{Shaded}

The ``NOTE'' parts address systematic errors we noticed and are only
relevant for detailed persona modeling and when including historical
context. In the general case, we deployed notes such as ``The text is in
an old language and may be erroneous due to technical processes.'' or
``In the past people wrote like they heard, so an entity is also valid
if it sounds like the real entity, even if written incorrectly''. All
notes ought to be adapted to the specific, observed errors of the used
{LLM} to improve performance.

\subsection{Matching spans through post-processing the
results}\label{matching-spans-through-post-processing-the-results}

Due to the heuristic nature and only approximative generation of {LLM}s,
the output is often not exactly reproducible. We especially noticed that
the {LLM} tends to correct ``mistakes'' in the source format or change
the hyphenation of entities. But this also enables the {LLM} to become
closer to a real annotator, who has similar freedoms. Due to this we
extracted the annotated spans and searched for them in the source
material with a fuzzy search provided by the python-library
\texttt{fuzzysearch}. We specifically used
\texttt{find\_near\_matches(span,\ text,\ len(span)\ //\ 5)} to allow
for up to 1 error every 5 generated characters, which corresponds to a
Levenshtein-Distance of {\textbf{n}} for a string at least length
{\textbf{5n}} (Levenshtein 1966). With this we had no errors in locating
the spans in the source-material.

For comparing if the same span is present in the generated annotation
and the ground truth, we employed the library \texttt{nervaluate}
(Batista and Upson {[}2019{]} 2020) to match the extracted spans. We
used the most lenient criteria for {NER} matching \texttt{ent\_type},
meaning it suffices to have an overlap with the annotation and having
the correct entity-type annotated (Segura-Bedmar, Mart{\'{\i}}nez, and
Herrero-Zazo 2013). This also yielded an average
precision/recall/F\textsubscript{1}-Score for the whole corpus as well
as every single page.

\subsection{Selection of LLMs}\label{selection-of-llms}

For this paper, we focussed on the \texttt{gpt4o-2024-08-06} model
(OpenAI 2024) with no retraining or more than a black-box access to the
{API.}\footnote{We set up everything as reproducible as possible -
  according to documentation. Although we assume no bad intent on
  OpenAI's part, all promises made could not be verified independently.
  On the notion of black box see Winner (1993).}

\subsection{Evaluating the performance of prompt
engineering}\label{evaluating-the-performance-of-prompt-engineering}

To measure the performance of our approach, we compare the results of
iterative prompts and their various components to two off-the-shelf
{NLP} libraries commonly used for {NER}: flair (Akbik et al. 2019) with
its \texttt{ner-german-large} model (Schweter and Akbik 2021), which has
a reported F\textsubscript{1}-Score of {\textbf{92.31}} on the CoNLL-03
German revised benchmark ({``State-of-the-{Art Models}''} {[}2018{]}
2025), and spaCy (Honnibal et al. 2020) using
\texttt{de\_core\_news\_lg}, for which spaCy claims a self-reported
F\textsubscript{1}-Score of {\textbf{85}} for the {NER}-Task after
training ({``De\_core\_news\_lg''} 2023).\footnote{SpaCy reports a
  F\textsubscript{1}-Score of 91.6 for the CoNLL-03 benchmark but used a
  {RoBERTa}-based transformer model specifically trained for {NER}.}

\subsubsection{Establishing baselines}\label{establishing-baselines}

Flair is straight-forward to implement. The German language model offers
the entity classes we needed (plus an additional \texttt{MISC}-class).
Results were also within our expectations across all classes. Due to the
model being trained on modern texts, we expected not to achieve the
reported F\textsubscript{1}-Score of 92 with historical sources (Our
corpus resulted in 0.76 recall; 0.89 precision; F\textsubscript{1}-Score
of 0.81, see tbl.~\ref{tbl:1}).

The baseline of spaCy on the other hand was surprisingly disappointing.
While having a recall of 0.82 in the \texttt{LOC}-Class and 0.76 in the
\texttt{PER}-Class, spaCy completely failed to recognize most of the
\texttt{ORG} (recall only 0.12). This is a known issue with
\texttt{ORG}s that are not in the original training-set are rarely
found. Additionally spaCy found 684 sequences as ``named entities'' that
were not annotated as such (while the whole corpus only had 1490
annotations in total) yielding an unsatisfactory precision. Over all
classes this only yielded mediocre results (Our corpus resulted in 0.59
recall; 0.44 precision; F\textsubscript{1}-Score of 0.50, see
tbl.~\ref{tbl:1}).

\section{Results}\label{results}

\subsection{Basic-Prompts}\label{basic-prompts}

We expected basic prompts without any context or more advanced
prompt-engineering to perform worse than a current large German model
for flair. But even with this simple approach, modern sophisticated
{LLM}s are able to at least perform on par with state-of-the-art
specialist {NLP} tools (recall 0.77 (ours) vs.~0.76 (flair); precision
0.91 (ours) vs.~0.89 (flair); F1 0.83 (ours) vs 0.81 (flair)).

This already yields good precision values. However, it is important to
emphasize that historians are particularly interested in high recall in
order to capture as many potential results as possible, which they can
then go through manually. Therefore, in the following test, we will
examine whether recall can be increased, especially by using contextual
information.

We achieve this comparison by analyzing three different dimensions of
optimization. First we change the persona-modeling aspect of the prompt
in two distinct ways: giving general guidance and giving source-specific
knowledge. Then we look at the improvement potential that additional
annotated examples would have

\subsection{Context (0-Shot)}\label{context-0-shot}

To analyze our main theory that context is core for unlocking the
highest performance we compared the 0-shot prompt with different
specifications of context.

\begin{itemize}
\tightlist
\item
  In the ``No Context''-case the instructions were merely information on
  how to annotate the data. No specific prompt-engineering techniques
  (like in the ablation study) were used.
\item
  The ``Generic Context'' takes a generic historic-persona (``As an
  experienced linguist specializing in Named Entity Recognition ({NER})
  for historical texts and expert in various historical epochs and
  locations, your task is to annotate a given text for machine
  reusability.'').
\item
  The ``Specific Context'' then also adds details on the specific
  material and what to expect (similar to the first paragraph of
  ``Corpus and data set'').
\item
  The ``Full Prompt'' integrates all the above, combined with several
  optimization techniques for prompting.
\item
  Finally, we tested the prompts in both German and English (mentioned
  in brackets below).
\item
  We report mean and standard deviation, as the annotation was done
  pagewise.
\end{itemize}

\begin{longtable}[]{@{}
  >{\raggedright\arraybackslash}p{(\columnwidth - 14\tabcolsep) * \real{0.1754}}
  >{\raggedright\arraybackslash}p{(\columnwidth - 14\tabcolsep) * \real{0.0351}}
  >{\raggedright\arraybackslash}p{(\columnwidth - 14\tabcolsep) * \real{0.1404}}
  >{\raggedright\arraybackslash}p{(\columnwidth - 14\tabcolsep) * \real{0.1228}}
  >{\raggedright\arraybackslash}p{(\columnwidth - 14\tabcolsep) * \real{0.1404}}
  >{\raggedright\arraybackslash}p{(\columnwidth - 14\tabcolsep) * \real{0.1228}}
  >{\raggedright\arraybackslash}p{(\columnwidth - 14\tabcolsep) * \real{0.1404}}
  >{\raggedright\arraybackslash}p{(\columnwidth - 14\tabcolsep) * \real{0.1228}}@{}}
\caption{\label{tbl:1}Different levels of context-specificity
({\textbf{without}} prompt-engineering techniques) compared to baseline
and full prompts ({\textbf{with}} prompt-engineering
techniques).}\tabularnewline
\toprule\noalign{}
\begin{minipage}[b]{\linewidth}\raggedright
\textbf{Context-Impact}
\end{minipage} & \begin{minipage}[b]{\linewidth}\raggedright
lang
\end{minipage} & \begin{minipage}[b]{\linewidth}\raggedright
\textbf{Recall}
\end{minipage} & \begin{minipage}[b]{\linewidth}\raggedright
\end{minipage} & \begin{minipage}[b]{\linewidth}\raggedright
\textbf{Precision}
\end{minipage} & \begin{minipage}[b]{\linewidth}\raggedright
\end{minipage} & \begin{minipage}[b]{\linewidth}\raggedright
\textbf{F\textsubscript{1}-Score}
\end{minipage} & \begin{minipage}[b]{\linewidth}\raggedright
\end{minipage} \\
\midrule\noalign{}
\endfirsthead
\toprule\noalign{}
\begin{minipage}[b]{\linewidth}\raggedright
\textbf{Context-Impact}
\end{minipage} & \begin{minipage}[b]{\linewidth}\raggedright
lang
\end{minipage} & \begin{minipage}[b]{\linewidth}\raggedright
\textbf{Recall}
\end{minipage} & \begin{minipage}[b]{\linewidth}\raggedright
\end{minipage} & \begin{minipage}[b]{\linewidth}\raggedright
\textbf{Precision}
\end{minipage} & \begin{minipage}[b]{\linewidth}\raggedright
\end{minipage} & \begin{minipage}[b]{\linewidth}\raggedright
\textbf{F\textsubscript{1}-Score}
\end{minipage} & \begin{minipage}[b]{\linewidth}\raggedright
\end{minipage} \\
\midrule\noalign{}
\endhead
\bottomrule\noalign{}
\endlastfoot
All 0-Shot & & \textbf{mean ± stdev} & \textbf{impact} & \textbf{mean ±
stdev} & \textbf{impact} & \textbf{mean ± stdev} & \textbf{impact} \\
Full Prompt & de & 0.84 ±0.10 & 0.00 \% & \textbf{0.91 ±0.08} &
\textbf{0.00 \%} & 0.87 ±0.08 & 0.00 \% \\
Full Prompt & en & 0.85 ±0.09 & 0.78 \% & 0.91 ±0.06 & -0.40 \% &
\textbf{0.88 ±0.07} & \textbf{+0.19 \%} \\
Specific Context & de & 0.81 ±0.19 & -3.58 \% & 0.87 ±0.19 & -3.84 \% &
0.84 ±0,19 & -3.66 \% \\
Specific Context & en & \textbf{0.86 ±0.08} & \textbf{+2.26 \%} & 0.89
±0.08 & -2.14 \% & 0.88 ±0.07 & +0.07 \% \\
Generic Context & de & 0.81 ±0.11 & -3.62 \% & 0.90 ±0.10 & -1.36 \% &
0.85 ±0.09 & -2.61 \% \\
No Context & de & 0.75 ±0.15 & -10.74 \% & 0.90 ±0.09 & -1.61 \% & 0.81
±0.11 & -7.17 \% \\
\textbf{\emph{Baseline flair}} & & \textbf{\emph{0.76 ±0.13}} &
\textbf{\emph{-9.98 \%}} & \textbf{\emph{0.89 ±0.10}} &
\textbf{\emph{-2.43 \%}} & \textbf{\emph{0.81 ±0.11}} &
\textbf{\emph{-7.38 \%}} \\
\textbf{\emph{Baseline spaCy}} & & \textbf{\emph{0.71 ±0.13}} &
\textbf{\emph{-15.91 \%}} & \textbf{\emph{0.62 ±0.11}} &
\textbf{\emph{-32.03 \%}} & \textbf{\emph{0.66 ±0.10}} &
\textbf{\emph{-24.53 \%}} \\
\end{longtable}

\subsection{Including examples: 1-shot, few-shot,
many-shot}\label{including-examples-1-shot-few-shot-many-shot}

Another way to improve the results and provide more context is through
one-shot or few-shot learning. Here, in addition to instructions in the
sense of the ``full prompt'' mentioned earlier, we also include examples
of the desired results. Brown et al. (2020) demonstrated that output
quality improves significantly when prompts integrate examples. We
consequently explore the impact of such examples. This approach allows
us to measure the impact of few-shots on the outcomes. At the same time,
it addresses the question of whether the added value of preparing such
examples is worthwhile at all.

Every example consists of small random excerpts of around 200-500
characters from the corpus that were not in the annotation set. This is
drastically lower than the whole page the model has to annotate in the
experiments, which are 4000-9000 characters long. However, each sample
needed to be manually annotated, making the process rather
labor-itensive. Nguyen and Wong (2023) have shown that prompts are
sensitive to choosing the {\emph{right}} examples. To control for the
impact of implicit domain knowledge, we chose random examples ensuring
an equal distribution of classes.

We incrementally increased the number of shots by powers of two from
2\^{}1 to a maximum of 2\^{}5. The maximum amount tested was 32-shot,
with all classes being balanced. In all other cases they were randomly
chosen once from those 32. This also has the side-effect of simulating
impacts from non-optimized examples as opposed to specifically
optimizing these for the given prompt with the risk of overfitting. Liu
et al. (2021) describe the impact of such optimizations in detail.

\begin{longtable}[]{@{}
  >{\raggedright\arraybackslash}p{(\columnwidth - 14\tabcolsep) * \real{0.1692}}
  >{\raggedright\arraybackslash}p{(\columnwidth - 14\tabcolsep) * \real{0.0462}}
  >{\raggedright\arraybackslash}p{(\columnwidth - 14\tabcolsep) * \real{0.1385}}
  >{\raggedright\arraybackslash}p{(\columnwidth - 14\tabcolsep) * \real{0.1231}}
  >{\raggedright\arraybackslash}p{(\columnwidth - 14\tabcolsep) * \real{0.1385}}
  >{\raggedright\arraybackslash}p{(\columnwidth - 14\tabcolsep) * \real{0.1231}}
  >{\raggedright\arraybackslash}p{(\columnwidth - 14\tabcolsep) * \real{0.1385}}
  >{\raggedright\arraybackslash}p{(\columnwidth - 14\tabcolsep) * \real{0.1231}}@{}}
\caption{\label{tbl:2}Impact of the number of examples given in addition
to specific context and all used prompt-engineering
techniques.}\tabularnewline
\toprule\noalign{}
\begin{minipage}[b]{\linewidth}\raggedright
\textbf{Shots}
\end{minipage} & \begin{minipage}[b]{\linewidth}\raggedright
lang
\end{minipage} & \begin{minipage}[b]{\linewidth}\raggedright
\textbf{Recall}
\end{minipage} & \begin{minipage}[b]{\linewidth}\raggedright
\end{minipage} & \begin{minipage}[b]{\linewidth}\raggedright
\textbf{Precision}
\end{minipage} & \begin{minipage}[b]{\linewidth}\raggedright
\end{minipage} & \begin{minipage}[b]{\linewidth}\raggedright
\textbf{F\textsubscript{1}-Score}
\end{minipage} & \begin{minipage}[b]{\linewidth}\raggedright
\end{minipage} \\
\midrule\noalign{}
\endfirsthead
\toprule\noalign{}
\begin{minipage}[b]{\linewidth}\raggedright
\textbf{Shots}
\end{minipage} & \begin{minipage}[b]{\linewidth}\raggedright
lang
\end{minipage} & \begin{minipage}[b]{\linewidth}\raggedright
\textbf{Recall}
\end{minipage} & \begin{minipage}[b]{\linewidth}\raggedright
\end{minipage} & \begin{minipage}[b]{\linewidth}\raggedright
\textbf{Precision}
\end{minipage} & \begin{minipage}[b]{\linewidth}\raggedright
\end{minipage} & \begin{minipage}[b]{\linewidth}\raggedright
\textbf{F\textsubscript{1}-Score}
\end{minipage} & \begin{minipage}[b]{\linewidth}\raggedright
\end{minipage} \\
\midrule\noalign{}
\endhead
\bottomrule\noalign{}
\endlastfoot
All Full Prompt & lang & \textbf{mean ± stdev} & \textbf{impact} &
\textbf{mean ± stdev} & \textbf{impact} & \textbf{mean ± stdev} &
\textbf{impact} \\
32 & de & \textbf{0.89 ±0.09} & \textbf{0.00 \%} & 0.90 ±0.06 & 0.00 \%
& \textbf{0.89 ±0.06} & \textbf{0.00 \%} \\
16 & de & 0.86 ±0.10 & -2.86 \% & 0.90 ±0.08 & -0.62 \% & 0.88 ±0.07 &
-1.81 \% \\
8 & en & 0.83 ±0.13 & -6.12 \% & 0.88 ±0.09 & -1.93 \% & 0.85 ±0.09 &
-4.40 \% \\
8 & de & 0.81 ±0.12 & -9.31 \% & 0.90 ±0.08 & 0.13 \% & 0.85 ±0.09 &
-5.15 \% \\
4 & de & 0.80 ±0.13 & -10.47 \% & 0.90 ±0.08 & 0.41 \% & 0.84 ±0.09 &
-5.82 \% \\
2 & de & 0.83 ±0.13 & -7.12 \% & 0.90 ±0.08 & 0.67 \% & 0.86 ±0.10 &
-3.65 \% \\
1 & en & 0.82 ±0.13 & -8.27 \% & 0.89 ±0.10 & -0.71 \% & 0.85 ±0.10 &
-4.83 \% \\
1 & de & 0.82 ±0.12 & -7.62 \% & \textbf{0.92 ±0.07} & \textbf{1.72 \%}
& 0.86 ±0.08 & -3.33 \% \\
0 & en & 0.85 ±0.09 & -4.50 \% & 0.91 ±0.06 & 1.11 \% & 0.88 ±0.07 &
-1.14 \% \\
0 & de & 0.84 ±0.10 & -5.02 \% & 0.91 ±0.08 & 1.27 \% & 0.87 ±0.08 &
-1.91 \% \\
\textbf{\emph{Baseline flair}} & & \textbf{\emph{0.76 ±0.13}} &
\textbf{\emph{-14.51 \%}} & \textbf{\emph{0.89 ±0.10}} &
\textbf{\emph{-1.20 \%}} & \textbf{\emph{0.81 ±0.11}} &
\textbf{\emph{-9.15 \%}} \\
\textbf{\emph{Baseline spaCy}} & & \textbf{\emph{0.71 ±0.13}} &
\textbf{\emph{-20.13 \%}} & \textbf{\emph{0.62 ±0.11}} &
\textbf{\emph{-31.17 \%}} & \textbf{\emph{0.66 ±0.10}} &
\textbf{\emph{-25.98 \%}} \\
\end{longtable}

The surprising result is that overall {\textbf{0-shot performs better
than few-shot}} in both recall and precision. Small improvements occur
only with 16 examples and more. {\textbf{We did not manage to hit a
ceiling}} at which recall stops improving - even with 32-shot. But we
were limited by the amount of tokens fitting inside the {LLM}s'
context-window, so that a further increase is not possible.

\subsection{Prompting language}\label{prompting-language}

Throughout our initial experiments we had prompted the {LLM}s in German
and expected a significant improvement of performance when switching to
English as the language for instructions and contextual information
(though not for the examples). This expectation was confirmed and
yielded a 5\% improvement in recall. The difference in precision was
much lower. This discrepancy can be nearly mitigated in German-language
prompts through prompt optimization techniques (tbl.~\ref{tbl:2}). To
our surprise we noticed that prompt-engineering is only relevant in a
meaningful way for German instructions, as the performance for English
instructions did not change considerably with or without
prompt-engineering. In consequence, and most importantly, the
differences between the performance of the two prompting languages using
our full prompts were insignificant. This will require further testing
with other languages but it certainly opens opportunities for a broader
application of our approach and to democratize access to such techniques
for non-English speaking communities.

\subsection{Ablation Study}\label{ablation-study}

In the ablation study, we enable and disable specific features of our
prompt in order to measure their impact on the overall outcome. The
ablation study was conducted on the German prompt only.

We first did the ablation study on the 32-shot prompt, but all results
were pretty similar due to the example-block dominating the whole
prompt. With so many examples using a multiple of tokens compared to all
instructions, the ablation-effect we want to measure here is basically
invisible. Therefore we switched to 0-shot for this analysis.

The parts added and reviewed individually were the following:

\begin{itemize}
\tightlist
\item
  {\textbf{only structure}}: addition of markdown-headings/separations
  of areas
\item
  {\textbf{only instruction-repetition}}: addition of the short summary
  of instructions at the end
\item
  {\textbf{only bullying}}: offering of reward/punishment for
  correct/wrong annotations
\item
  {\textbf{only system-prompt}}: providing the input as two distinct
  messages marked ``system'' containing the prompt and ``user''
  containing the data versus putting everything in one ``user''-message
\end{itemize}

For comparison we added selected results from the previous tables
regarding different context levels (in italics) as well as the baseline
results. ``PE'' here refers to using all techniques, otherwise none
other than the mentioned one was used. As zero for our benchmark we set
a prompt with specific contextual information but no prompt-engineering
techniques.

\begin{longtable}[]{@{}
  >{\raggedright\arraybackslash}p{(\columnwidth - 12\tabcolsep) * \real{0.3130}}
  >{\raggedright\arraybackslash}p{(\columnwidth - 12\tabcolsep) * \real{0.1221}}
  >{\raggedright\arraybackslash}p{(\columnwidth - 12\tabcolsep) * \real{0.1069}}
  >{\raggedright\arraybackslash}p{(\columnwidth - 12\tabcolsep) * \real{0.1221}}
  >{\raggedright\arraybackslash}p{(\columnwidth - 12\tabcolsep) * \real{0.1069}}
  >{\raggedright\arraybackslash}p{(\columnwidth - 12\tabcolsep) * \real{0.1221}}
  >{\raggedright\arraybackslash}p{(\columnwidth - 12\tabcolsep) * \real{0.1069}}@{}}
\caption{\label{tbl:3}Ablation study with impact of different
prompt-engineering techniques on the overall recognition; gray/italic
results are from previous tables for comparison.}\tabularnewline
\toprule\noalign{}
\begin{minipage}[b]{\linewidth}\raggedright
\textbf{Ablation}
\end{minipage} & \begin{minipage}[b]{\linewidth}\raggedright
\textbf{Recall}
\end{minipage} & \begin{minipage}[b]{\linewidth}\raggedright
\end{minipage} & \begin{minipage}[b]{\linewidth}\raggedright
\textbf{Precision}
\end{minipage} & \begin{minipage}[b]{\linewidth}\raggedright
\end{minipage} & \begin{minipage}[b]{\linewidth}\raggedright
\textbf{F\textsubscript{1}-Score}
\end{minipage} & \begin{minipage}[b]{\linewidth}\raggedright
\end{minipage} \\
\midrule\noalign{}
\endfirsthead
\toprule\noalign{}
\begin{minipage}[b]{\linewidth}\raggedright
\textbf{Ablation}
\end{minipage} & \begin{minipage}[b]{\linewidth}\raggedright
\textbf{Recall}
\end{minipage} & \begin{minipage}[b]{\linewidth}\raggedright
\end{minipage} & \begin{minipage}[b]{\linewidth}\raggedright
\textbf{Precision}
\end{minipage} & \begin{minipage}[b]{\linewidth}\raggedright
\end{minipage} & \begin{minipage}[b]{\linewidth}\raggedright
\textbf{F\textsubscript{1}-Score}
\end{minipage} & \begin{minipage}[b]{\linewidth}\raggedright
\end{minipage} \\
\midrule\noalign{}
\endhead
\bottomrule\noalign{}
\endlastfoot
All 0-Shot & \textbf{mean ± stdev} & \textbf{impact} & \textbf{mean ±
stdev} & \textbf{impact} & \textbf{mean ± stdev} & \textbf{impact} \\
\emph{Specific Context + PE} & \emph{0.84 ±0.10} & \emph{3.71 \%} &
\textbf{\emph{0.91 ±0.08}} & \textbf{\emph{3.99 \%}} & \emph{0.87 ±0.08}
& \emph{3.80 \%} \\
Specific Context + structure & \textbf{0.85 ±0.09} & \textbf{5.10 \%} &
0.91 ±0.08 & 3.84 \% & \textbf{0.88 ±0.08} & \textbf{4.40 \%} \\
Specific Context + system-prompt & 0.85 ±0.10 & 4.04 \% & 0.91 ±0.08 &
3.88 \% & 0.88 ±0.08 & 3.95 \% \\
Specific Context + instruction-repetition & 0.82 ±0.19 & 0.26 \% & 0.88
±0.19 & 0.28 \% & 0.85 ±0.19 & 0.31 \% \\
Specific Context + bullying & 0.83 ±0.20 & 1.44 \% & 0.87 ±0.20 & -0.82
\% & 0.84 ±0.19 & 0.20 \% \\
\emph{Specific Context} & \emph{0.81 ±0.19} & 0.00 \% & \emph{0.87
±0.19} & 0.00 \% & \emph{0.84 ±0.19} & 0.00 \% \\
\emph{Generic Context + PE} & \emph{0.80 ±0.11} & \emph{-2.35 \%} &
\emph{0.92 ±0.10} & \emph{2.92 \%} & \emph{0.84 ±0.10} & \emph{-0.07
\%} \\
\emph{Generic Context} & \emph{0.81 ±0.11} & \emph{-0.05 \%} &
\emph{0.90 ±0.10} & \emph{2.58 \%} & \emph{0.85 ±0.09} & \emph{1.09
\%} \\
\emph{No Context + PE} & \emph{0.74 ±0.15} & \emph{-8.59 \%} &
\emph{0.91 ±0.10} & \emph{3.63 \%} & \emph{0.81 ±0.11} & \emph{-3.66
\%} \\
\emph{No Context} & \emph{0.75 ±0.15} & \emph{-7.43 \%} & \emph{0.90
±0.09} & \emph{2.31 \%} & \emph{0.81 ±0.11} & \emph{-3.64 \%} \\
\textbf{\emph{Baseline flair}} & \textbf{\emph{0.76 ±0.13}} &
\textbf{\emph{-6.65 \%}} & \textbf{\emph{0.89 ±0.10}} &
\textbf{\emph{1.46 \%}} & \textbf{\emph{0.81 ±0.11}} &
\textbf{\emph{-3.86 \%}} \\
\textbf{\emph{Baseline spaCy}} & \textbf{\emph{0.71 ±0.13}} &
\textbf{\emph{-12.79 \%}} & \textbf{\emph{0.62 ±0.11}} &
\textbf{\emph{-29.32 \%}} & \textbf{\emph{0.66 ±0.10}} &
\textbf{\emph{-21.66 \%}} \\
\end{longtable}

We could already show that context is an important driver for achieving
great performance. But without prompt-engineering techniques the
difference between generic and specific context is within each other's
standard-deviations. Adding well-known {\textbf{prompt-engineering
techniques however, seems to behave orthogonal to contextual clues}}, as
only adding all techniques does not improve the results (``No Context''
vs.~``No Context + PE'').

\section{Discussion}\label{discussion}

Our results show that while advanced prompting strategies improve
scores, their impact is limited, typically increasing performance by
only 1-2\% or (as in our case) negating a difference in the instruction
language. Similarly, the number of examples in few-shot learning has
shown some more substantial impact only if there are 16 or more shots,
while creating appropriate examples is labor-intensive and requires
expert knowledge to avoid bias and ensure comprehensive coverage.

In contrast, providing explicit and detailed context has a much greater
impact. Just providing general context, as has been the norm, improves
recall by 6\%. However, when specific, detailed context is provided, it
leads to further, significant improvement of the results by another 5\%.
Which in the end, sets our approach in terms of retrieval 10\% above
flair and 15\% above spaCy, in terms of F\textsubscript{1}-Score 7\%
above flair and 22\% above spaCy.

The results clearly demonstrate that instead of relying solely on the
model's capabilities as a language model in handling language ({NLP}),
the outcomes improve significantly when the task is defined as a
content-based one, thereby incorporating the domain knowledge
represented within the model through appropriate prompting---while
acknowledging that this represents only a limited perspective of the
world. In doing so, the model can surpass human capabilities in certain
areas.

To cite an example for this: when annotating the sequence ``Borstells
Lesezirkel,'' human annotators labeled ``Borstells'' as a person
(\texttt{PER}), whereas the model correctly identified the entire phrase
as an organization (\texttt{ORG}). Further investigation confirmed that
``Borstells Lesezirkel'' refers to a commercial lending library, a fact
unknown to the human annotators but represented within the model's
knowledge base. When the model is separately asked about ``Borstells
Lesezirkel,'' it provides additional information, which, while not
entirely accurate, aids human editors in correctly determining the
category. This opens avenues for future work on iterative generation of
ground truth with the help of {LLM}s.

On the other hand, the result cannot be attributed to a more flexible
language understanding of the model. In our experiment, when no
contextual information was provided, the model achieved the exact same
F\textsubscript{1}-Score of 81\% as the more universally trained Flair
model. Both outperformed the spaCy model, which was trained on newspaper
texts from the 1990s.

This demonstrates that well-formulated prompts including much more
specific information of the task at hand can effectively leverage the
model's representation of knowledge in this domain and thus enhance, in
our case, detection and classification. Within the constraints of its
limitations regarding bias and representation, as well as its dependence
on training data, the model operates here akin to an omniscient
historian, capable of outperforming human annotators within the scope of
its represented knowledge.

Furthermore, we could show that prompt-engineering techniques evened out
differences in the performances of German and English prompts, while the
latter outperformed the former without such approaches.

Our ablation study indicates that while prompt engineering strategies
individually contribute to enhanced {LLM} performance in {NER}, their
effectiveness is markedly amplified when paired with contextually rich
information. However, it's noteworthy that there isn't a straightforward
additive relationship between the two; instead, they intertwine, each
influencing the efficacy of the other. Rather than merely stacking
benefits from additional prompt examples or refining structure alone,
their synergy seems to unlock nuanced understanding within {LLM}s,
leading to better recognition outcomes without guaranteeing perfection.
The intricacies of historical texts demand a delicate balance between
explicit instructions and embedded contextual knowledge---a combination
that requires more exploration for its potential full realization in
practical applications. This finding encourages a tempered enthusiasm as
we continue to refine our approach, acknowledging the complexity of
language models' interaction with humanities-informed data.

\subsection{Transferability and future
work}\label{transferability-and-future-work}

In terms of transferability, preliminary tests on documents from the
16\textsuperscript{th} to the 18\textsuperscript{th} centuries from a
variety of genres suggest that our methodology can be applied to texts
from different historical periods and backgrounds with clear
improvements over the respective baselines of Flair and spaCy, although
further testing is needed to fully understand the scope of applicability
of this approach to a wider range of historical documents and use cases.

In particular, since our approach may also provide a way to annotate
unique classes without first having to manually generate sufficient
training data and perform fine-tuning, simply by describing the class in
natural language. Having established in this paper the importance of
detailed and task-specific contextual information, future work will
explore its transferability to other, more specific tasks in the field
of historical research. Future work will include:

\begin{itemize}
\tightlist
\item
  Investigating how well our approach can handle earlier linguistic
  forms to determine its broader applicability across different
  historical periods and languages,
\item
  Experimenting with unique, case-specific classes that have not been
  used before, which would normally require specific training data and
  fine-tuning, but could be solved based on our approach by a
  well-thought-through natural description of the class,
\item
  Investigate the extent to which a model can find the specifics of a
  given text itself and thus improve the quality of the contextual
  information, or even create it by itself, thereby making more
  extensive use of the content-based capabilities of the models,
\item
  how the different models and their different capacities can influence
  their performance in these tasks, and how more specific context
  information is related to different prompt engineering techniques,
\item
  and finally, to investigate the extent to which limitations in the
  results are due to ambiguities inherent in the historical texts, and
  how we can deal with these ambiguities and represent them in the
  evaluation.
\end{itemize}

\section{Conclusion}\label{conclusion}

In our study, we have explored the use of {LLM}s for {NER} in historical
and low-resource texts through humanities-informed approaches,
demonstrating a significant improvement in performance compared to
traditional methods as they are implemented state-of-the-art
{NLP}-frameworks like flair or spaCy. The results indicate that
incorporating specific contextual information into prompts is
fundamental for achieving high accuracy in {NER} tasks.

The integration of domain knowledge through humanities-informed
prompting has shown to significantly enhance {LLM} performance in {NER}
tasks for historical texts. This method not only facilitates more
efficient analysis but also brings us closer to replicating the
expertise traditionally required by human annotators, which could
revolutionize how we approach textual analysis within digital humanities
and beyond.

\section{Acknowledgements}\label{acknowledgements}

\subsection{Use of artificial intelligence (AI)
tools}\label{use-of-artificial-intelligence-ai-tools}

As non-native speakers of English, some of our co-authors edited parts
of their contributions for linguistic and idiomatic clarity. To this
end, draft passages were submitted to ChatGPT-4o (OpenAI 2024) and
\href{https://www.deepl.com/en/write}{DeepL Write}. During this process,
all suggested changes were individually verified and accepted or
rejected accordingly. The text was then finalised and edited without
further resort to {AI} tools.

\subsection{Funding disclosure
statement}\label{funding-disclosure-statement}

Parts of the research have been funded through the German National
Research Data Infrastructure ({NFDI}) consortium 4Memory
(\textless{}\href{http://www.4memory.de}{www.4memory.de}\textgreater)
and the {AI-Skills}
(\textless{}\href{http://www.ai-skills.hu-berlin.de}{www.ai-skills.hu-berlin.de}\textgreater)
project at Humboldt-Universität zu Berlin. We gratefully acknowledge the
financial support of the German Research Foundation ({DFG}) (4Memory,
project no. 501609550), the Federal Ministry of Education and Research
({BMBF}) ({AI-Skills}, project funding code 16DHBKI014) and the Berlin
Senate Department for Science, Health and Care ({AI-Skills}).

\subsection{Authorship contribution statement
(CRediT)}\label{authorship-contribution-statement-credit}

\begin{itemize}
\tightlist
\item
  Conceptualization: Torsten Hiltmann, Martin Dröge
\item
  Data curation: Martin Dröge, Wiebke Sczeponik, Koray Mendi, Paul
  Bayer, Anica Skibba, Philipp Schneider, Sophie Eckenstaler
\item
  Formal analysis: Nicole Dresselhaus
\item
  Methodology: Torsten Hiltmann, Martin Dröge, Nicole Dresselhaus
\item
  Software: Nicole Dresselhaus
\item
  Writing - original draft: Torsten Hiltmann, Nicole Dresselhaus, Martin
  Dröge, Till Grallert, Jascha Schmitz, Melanie Althage, Philipp
  Schneider
\item
  Writing - review \& editing: Till Grallert
\end{itemize}

\section{Data~availability statement}\label{data-availability-statement}

All data and code is available under free and open licenses and
\texttt{\textless{}anonymized\textgreater{}} at
\textless{}\url{https://osf.io/yr5ck/?view_only=a9b2b10af0f247048d64b2bf4415b5c8}\textgreater.

\section*{Bibliography}\label{bibliography}
\addcontentsline{toc}{section}{Bibliography}

\phantomsection\label{refs}


\begin{CSLReferences}{1}{0}
\bibitem[\citeproctext]{ref-AkbikEtAl2019FLAIREasytoUse}
Akbik, Alan, Tanja Bergmann, Duncan Blythe, Kashif Rasul, Stefan
Schweter, and Roland Vollgraf. 2019. {``{FLAIR}: {An Easy-to-Use
Framework} for {State-of-the-Art NLP}.''} In \emph{Proceedings of the
2019 {Conference} of the {North American Chapter} of the {Association}
for {Computational Linguistics} ({Demonstrations})}, 54--59.
Minneapolis: Association for Computational Linguistics.
\url{https://doi.org/10.18653/v1/N19-4010}.

\bibitem[\citeproctext]{ref-BaedekerGraupe1921BerlinundUmgebung}
Baedeker, Karl, and Bruno Graupe. 1921. \emph{Berlin Und {Umgebung}:
{Handbuch} Für {Reisende}}. 19th ed. Leipzig: Karl Baedeker.
\url{https://nbn-resolving.org/urn:nbn:de:kobv:11-717582}.

\bibitem[\citeproctext]{ref-BatistaUpson2020nervaluate}
Batista, David, and Matthew Antony Upson. (2019) 2020. {``Nervaluate.''}
\url{https://github.com/mantisnlp/nervaluate}.

\bibitem[\citeproctext]{ref-BrownEtAl2020LanguageModels}
Brown, Tom B., Benjamin Mann, Nick Ryder, Melanie Subbiah, Jared Kaplan,
Prafulla Dhariwal, Arvind Neelakantan, et al. 2020. {``Language {Models}
Are {Few-Shot Learners}.''} July 22, 2020.
\url{https://doi.org/10.48550/arXiv.2005.14165}.

\bibitem[\citeproctext]{ref-BsharatEtAl2024PrincipledInstructions}
Bsharat, Sondos Mahmoud, Aidar Myrzakhan, and Zhiqiang Shen. 2024.
{``Principled {Instructions Are All You Need} for {Questioning
LLaMA-1}/2, {GPT-3}.5/4.''} January 18, 2024.
\url{http://arxiv.org/abs/2312.16171}.

\bibitem[\citeproctext]{ref-ChenEtAl2023LearningcontextLearning}
Chen, Jiawei, Yaojie Lu, Hongyu Lin, Jie Lou, Wei Jia, Dai Dai, Hua Wu,
Boxi Cao, Xianpei Han, and Le Sun. 2023. {``Learning {In-context
Learning} for {Named Entity Recognition}.''} May 26, 2023.
\url{https://doi.org/10.48550/arXiv.2305.11038}.

\bibitem[\citeproctext]{ref-ChenEtAl2015studyactivelearning}
Chen, Yukun, Thomas A. Lasko, Qiaozhu Mei, Joshua C. Denny, and Hua Xu.
2015. {``A Study of Active Learning Methods for Named Entity Recognition
in Clinical Text.''} \emph{Journal of Biomedical Informatics} 58
(December): 11--18. \url{https://doi.org/10.1016/j.jbi.2015.09.010}.

\bibitem[\citeproctext]{ref-DeToniEtAl2022EntitiesDates}
De Toni, Francesco, Christopher Akiki, Javier de la Rosa, Clémentine
Fourrier, Enrique Manjavacas, Stefan Schweter, and Daniel van Strien.
2022. {``Entities, {Dates}, and {Languages}: {Zero-Shot} on {Historical
Texts} with {T0}.''} \url{http://arxiv.org/abs/2204.05211}.

\bibitem[\citeproctext]{ref-spacy2023De_core_news_lg}
{``De\_core\_news\_lg.''} 2023. Explosion.
\url{https://github.com/explosion/spacy-models/releases/tag/de_core_news_lg-3.7.0}.

\bibitem[\citeproctext]{ref-Ehrmann2008EntiteesNommees}
Ehrmann, Maud. 2008. {``Les Entitées Nommées, de la linguistique au TAL
: Statut théorique et méthodes de désambiguïsation.''} PhD thesis, Paris
Diderot University. \url{https://hal.science/tel-01639190}.

\bibitem[\citeproctext]{ref-EhrmannEtAl2023NamedEntityRecognition}
Ehrmann, Maud, Ahmed Hamdi, Elvys Linhares Pontes, Matteo Romanello, and
Antoine Doucet. 2023. {``Named {Entity Recognition} and {Classification}
in {Historical Documents}: {A Survey}.''} \emph{ACM Computing Surveys}
56 (2): 27:1--47. \url{https://doi.org/10.1145/3604931}.

\bibitem[\citeproctext]{ref-EhrmannEtAl2020ImpressoNamedEntity}
Ehrmann, Maud, Camille Watter, Matteo Romanello, Simon Clematide, and
Flückiger. 2020. {``Impresso {Named Entity Annotation Guidelines},''}
January. \url{https://doi.org/10.5281/zenodo.3604227}.

\bibitem[\citeproctext]{ref-VanEijnatten2024Dutchintellectualculture}
Eijnatten, Jorsi van. 2024. {``Dutch Intellectual Culture Between 1962
and 1995, or, Using Classical Algorithms and {LLMs} to Efficiently
Extract Data with Imperfect {OCR}.''} In.
\url{https://2024.dhbenelux.org/wp-content/uploads/2024/05/DHB24_paper_van_Eijnatten_Dutch-intellectual-culture-between-1962-and-1995.pdf}.

\bibitem[\citeproctext]{ref-FederiakinEtAl2024Promptengineeringnew}
Federiakin, Denis, Dimitri Molerov, Olga Zlatkin-Troitschanskaia, and
Andreas Maur. 2024. {``Prompt Engineering as a New 21st Century
Skill.''} \emph{Frontiers in Education} 9 (November): 1366434.
\url{https://doi.org/10.3389/feduc.2024.1366434}.

\bibitem[\citeproctext]{ref-GiovanelliTraviglia2024AIKoGAMAIdriven}
Giovanelli, Riccardo, and Arianna Traviglia. 2024. {``{AIKoGAM}: {An
AI-driven Knowledge Graph} of the {Antiquities Market}: {Toward
Automatised Methods} to {Identify Illicit Trafficking Networks}.''}
\emph{Journal of Computer Applications in Archaeology} 7 (January):
92--114. \url{https://doi.org/gtr3s6}.

\bibitem[\citeproctext]{ref-GonzalezGallardoEtAl2023YesCanChatGPT}
González-Gallardo, Carlos-Emiliano, Emanuela Boros, Nancy Girdhar, Ahmed
Hamdi, Jose G. Moreno, and Antoine Doucet. 2023. {``Yes but.. {Can
ChatGPT Identify Entities} in {Historical Documents}?''} \emph{2023
ACM/IEEE Joint Conference on Digital Libraries (JCDL)}, June, 184--89.
\url{https://doi.org/10.1109/JCDL57899.2023.00034}.

\bibitem[\citeproctext]{ref-GrahamYatesInvestigatingAntiquities}
Graham, Shawn, Donna Yates, and Ahmed El-Roby. 2023. {``Investigating
Antiquities Trafficking with Generative Pre-Trained Transformer
({GPT})-3 Enabled Knowledge Graphs: {A} Case Study.''} \emph{Open
Research Europe} 3 (100). \url{https://doi.org/gtbsdp}.

\bibitem[\citeproctext]{ref-HollaEtAl2024LargeLanguageModels}
Holla, Kiran Voderhobli, Chaithanya Kumar, and Aryan Singh. 2024.
{``Large {Language Models} Aren't All That You Need.''}
\url{https://doi.org/10.48550/ARXIV.2401.00698}.

\bibitem[\citeproctext]{ref-HonnibalEtAl2020spaCy}
Honnibal, Matthew, Ines Montani, Sofie Van Landeghem, and Adriane Boyd.
2020. {``{spaCy}: {Industrial-strength} Natural Language Processing in
Python.''} \url{https://doi.org/10.5281/zenodo.1212303}.

\bibitem[\citeproctext]{ref-Karjus2023Machineassistedmixedmethods}
Karjus, Andres. 2023. {``Machine-Assisted Mixed Methods: Augmenting
Humanities and Social Sciences with Artificial Intelligence.''}
September 24, 2023. \url{https://doi.org/10.48550/arXiv.2309.14379}.

\bibitem[\citeproctext]{ref-KeraghelEtAl2024SurveyRecent}
Keraghel, Imed, Stanislas Morbieu, and Mohamed Nadif. 2024. {``A Survey
on Recent Advances in Named Entity Recognition.''} January 19, 2024.
\url{https://doi.org/10.48550/arXiv.2401.10825}.

\bibitem[\citeproctext]{ref-KlieEtAl2018INCEpTION}
Klie, Jan-Christoph, Michael Bugert, Beto Boullosa, Richard Eckart de
Castilho, and Iryna Gurevych. 2018. {``The {INCEpTION} Platform:
{Machine-assisted} and Knowledge-Oriented Interactive Annotation.''} In
\emph{Proceedings of the 27th International Conference on Computational
Linguistics: {System} Demonstrations}, 5--9. Santa Fe: Association for
Computational Linguistics.
\url{http://tubiblio.ulb.tu-darmstadt.de/106270/}.

\bibitem[\citeproctext]{ref-Levenshtein1966BinaryCodesCapable}
Levenshtein, V. I. 1966. {``Binary {Codes Capable} of {Correcting
Deletions}, {Insertions} and {Reversals}.''} \emph{Soviet Physics
Doklady} 10 (February): 707.
\url{https://ui.adsabs.harvard.edu/abs/1966SPhD...10..707L}.

\bibitem[\citeproctext]{ref-LevyEtAl2023DiverseDemonstrations}
Levy, Itay, Ben Bogin, and Jonathan Berant. 2023. {``Diverse
{Demonstrations Improve In-context Compositional Generalization}.''} In
\emph{Proceedings of the 61st {Annual Meeting} of the {Association} for
{Computational Linguistics} ({Volume} 1: {Long Papers})}, 1401--22.
Toronto, Canada: Association for Computational Linguistics.
\url{https://doi.org/10.18653/v1/2023.acl-long.78}.

\bibitem[\citeproctext]{ref-LiuEtAl2021WhatMakesGood}
Liu, Jiachang, Dinghan Shen, Yizhe Zhang, Bill Dolan, Lawrence Carin,
and Weizhu Chen. 2021. {``What {Makes Good In-Context Examples} for
{GPT-3}?''} January 17, 2021. \url{http://arxiv.org/abs/2101.06804}.

\bibitem[\citeproctext]{ref-NguyenWong2023IncontextExampleSelection}
Nguyen, Tai, and Eric Wong. 2023. {``In-Context {Example Selection} with
{Influences}.''} June 5, 2023.
\url{https://doi.org/10.48550/arXiv.2302.11042}.

\bibitem[\citeproctext]{ref-OpenAI2023ChatGPTLargelanguage}
OpenAI. 2022. {``{ChatGPT} {[}{Large} Language Model{]}.''}
\url{https://openai.com/index/chatgpt/}.

\bibitem[\citeproctext]{ref-OpenAIGPT4o}
---------. 2024. {``{ChatGPT-4o}.''}
\url{https://openai.com/index/hello-gpt-4o/}.

\bibitem[\citeproctext]{ref-OpenAIEtAl2024GPT4TechnicalReport}
OpenAI, Josh Achiam, Steven Adler, Sandhini Agarwal, Lama Ahmad, Ilge
Akkaya, Florencia Leoni Aleman, et al. 2024. {``{GPT-4 Technical
Report}.''} March 4, 2024. \url{http://arxiv.org/abs/2303.08774}.

\bibitem[\citeproctext]{ref-Pakhale2023ComprehensiveOverviewNamed}
Pakhale, Kalyani. 2023. {``Comprehensive {Overview} of {Named Entity
Recognition}: {Models}, {Domain-Specific Applications} and
{Challenges}.''} September 25, 2023.
\url{https://doi.org/10.48550/arXiv.2309.14084}.

\bibitem[\citeproctext]{ref-ReulEtAl2019OCR4allOpenSource}
Reul, Christian, Dennis Christ, Alexander Hartelt, Nico Balbach,
Maximilian Wehner, Uwe Springmann, Christoph Wick, Christine Grundig,
Andreas Büttner, and Frank Puppe. 2019. {``{OCR4all}: {An Open-Source
Tool Providing} a ({Semi-}){Automatic OCR Workflow} for {Historical
Printings}.''} \emph{Applied Sciences} 9 (22, 22).
\url{https://doi.org/10.3390/app9224853}.

\bibitem[\citeproctext]{ref-CalamariModels2021v2}
Reul, Christian, and Christoph Wick. 2021.
{``Calamari-{OCR}/Calamari\_models: {Pretrained} Mixed Models to Be Used
with {Calamari}.''}
\url{https://github.com/Calamari-OCR/calamari_models}.

\bibitem[\citeproctext]{ref-Rosenzweig2003ScarcityAbundancePreserving}
Rosenzweig, Roy. 2003. {``Scarcity or {Abundance}? {Preserving} the
{Past} in a {Digital Era}.''} \emph{The American Historical Review} 108
(3): 735--62. \url{https://doi.org/10.1086/ahr/108.3.735}.

\bibitem[\citeproctext]{ref-SanhEtAl2022MultitaskPromptedTraining}
Sanh, Victor, Albert Webson, Colin Raffel, Stephen H. Bach, Lintang
Sutawika, Zaid Alyafeai, Antoine Chaffin, et al. 2022. {``Multitask
{Prompted Training Enables Zero-Shot Task Generalization}.''} March 17,
2022. \url{https://doi.org/10.48550/arXiv.2110.08207}.

\bibitem[\citeproctext]{ref-SantosEtAl2024Namedentityrecognition}
Santos, Joaquim, Helena Freire Cameron, F. Olival, Fátima Farrica, and
Renata Vieira. 2024. {``Named Entity Recognition Specialised for
{Portuguese} 18th-Century {History} Research.''} In.
\url{https://www.semanticscholar.org/paper/Named-entity-recognition-specialised-for-Portuguese-Santos-Cameron/b1efd28f5ae7faeaff6e01a640497ac15d9d8028}.

\bibitem[\citeproctext]{ref-SarkerEtAl2024SeventeenthCenturySpanishAmerican}
Sarker, Shraboni, Ahmad Tamim Hamad, Hulayyil Alshammari, Viviana
Grieco, and Praveen Rao. 2024. {``Seventeenth-{Century Spanish American
Notary Records} for {Fine-Tuning Spanish Large Language Models}.''} June
9, 2024. \url{https://doi.org/10.48550/arXiv.2406.05812}.

\bibitem[\citeproctext]{ref-SchweterAkbik2021FLERTDocumentLevelFeatures}
Schweter, Stefan, and Alan Akbik. 2021. {``{FLERT}: {Document-Level
Features} for {Named Entity Recognition}.''} May 14, 2021.
\url{https://doi.org/10.48550/arXiv.2011.06993}.

\bibitem[\citeproctext]{ref-SeguraBedmarEtAl2013SemEval2013task9}
Segura-Bedmar, Isabel, Paloma Mart{\'{\i}}nez, and Mar{\'{\i}}a Herrero-Zazo. 2013.
{``{SemEval-2013} Task 9: {Extraction} of Drug-Drug Interactions from
Biomedical Texts ({DDIExtraction} 2013).''} In \emph{Second Joint
Conference on Lexical and Computational Semantics (*{SEM}), Volume 2:
{Proceedings} of the Seventh International Workshop on Semantic
Evaluation ({SemEval} 2013)}, edited by Suresh Manandhar and Deniz
Yuret, 341--50. Atlanta, Georgia, USA: Association for Computational
Linguistics. \url{https://aclanthology.org/S13-2056}.

\bibitem[\citeproctext]{ref-StammbachEtAl2022HeroesVillains}
Stammbach, Dominik, Maria Antoniak, and Elliott Ash. 2022. {``Heroes,
{Villains}, and {Victims}, and {GPT-3}: {Automated Extraction} of
{Character Roles Without Training Data}.''} In \emph{Proceedings of the
4th {Workshop} of {Narrative Understanding} ({WNU2022})}, 47--56.
Seattle, United States: Association for Computational Linguistics.
\url{https://doi.org/10.18653/v1/2022.wnu-1.6}.

\bibitem[\citeproctext]{ref-flairNLPflair2025}
{``State-of-the-{Art Models}.''} (2018) 2025. flairNLP/flair. January
28, 2025. \url{https://github.com/flairNLP/flair}.

\bibitem[\citeproctext]{ref-TangEtAl2024CHisIECInformationExtraction}
Tang, Xuemei, Zekun Deng, Qi Su, Hao Yang, and Jun Wang. 2024.
{``{CHisIEC}: {An Information Extraction Corpus} for {Ancient Chinese
History}.''} April 20, 2024.
\url{https://doi.org/10.48550/arXiv.2403.15088}.

\bibitem[\citeproctext]{ref-TaoriEtAl2023AlpacaStrongReplicable}
Taori, Rohan, Ishaan Gulrajani, Tianyi Zhang, Yann Dubois, Li Xuechen,
Carlos Guestrin, Percy Liang, and Tatsunori B. Hasimoto. 2023.
{``Alpaca: {A Strong}, {Replicable Instruction-Following Model}.''}
\url{https://crfm.stanford.edu/2023/03/13/alpaca.html}.

\bibitem[\citeproctext]{ref-ThomasEtAl2024LeveragingLLMsPostOCR}
Thomas, Alan, Robert Gaizauskas, and Haiping Lu. 2024. {``Leveraging
{LLMs} for {Post-OCR Correction} of {Historical Newspapers}.''} In
\emph{Proceedings of the {Third Workshop} on {Language Technologies} for
{Historical} and {Ancient Languages} ({LT4HALA}) @ {LREC-COLING-2024}},
edited by Rachele Sprugnoli and Marco Passarotti, 116--21. Torino,
Italia: {ELRA and ICCL}.
\url{https://aclanthology.org/2024.lt4hala-1.14}.

\bibitem[\citeproctext]{ref-THUDM2023ChatGLM26B}
THUDM. 2023. {``{ChatGLM2-6B}.''} THUKEG.
\url{https://github.com/THUDM/ChatGLM2-6B/blob/main/README_EN.md}.

\bibitem[\citeproctext]{ref-TouvronEtAl2023LlamaOpenFoundation}
Touvron, Hugo, Louis Martin, and Kevin Stone. 2023. {``Llama 2: {Open
Foundation} and {Fine-Tuned Chat Models}.''}

\bibitem[\citeproctext]{ref-WangEtAl2023GPTNERNamedEntity}
Wang, Shuhe, Xiaofei Sun, Xiaoya Li, Rongbin Ouyang, Fei Wu, Tianwei
Zhang, Jiwei Li, and Guoyin Wang. 2023. {``{GPT-NER}: {Named Entity
Recognition} via {Large Language Models}.''} October 7, 2023.
\url{https://doi.org/10.48550/arXiv.2304.10428}.

\bibitem[\citeproctext]{ref-WickEtAl2020CalamariHighPerformance}
Wick, Christoph, Christian Reul, and Frank Puppe. 2020. {``Calamari – {A
High-Performance Tensorflow-based Deep Learning Package} for {Optical
Character Recognition}.''} \emph{Digital Humanities Quarterly} 14 (2).
\url{https://arxiv.org/abs/1807.02004}.

\bibitem[\citeproctext]{ref-Winner1993OpeningBlack}
Winner, Langdon. 1993. {``Upon Opening the Black Box and Finding It
Empty: {Social} Constructivism and the Philosophy of Technology.''}
\emph{Science, Technology, \& Human Values} 18 (3, 3): 362--78.
\url{https://doi.org/10.1177/016224399301800306}.

\bibitem[\citeproctext]{ref-YanEtAl2024LTNERLargeLanguage}
Yan, Faren, Peng Yu, and Xin Chen. 2024. {``{LTNER}: {Large Language
Model Tagging} for {Named Entity Recognition} with {Contextualized
Entity Marking}.''} April 8, 2024.
\url{https://doi.org/10.48550/arXiv.2404.05624}.

\bibitem[\citeproctext]{ref-ZixiaoEtAl2024MICLImprovingContext}
Zixiao, Zhu, Feng Zijian, Zhou Hanzhang, Qian Junlang, and Mao Kezhi.
2024. {``{MICL}: {Improving In-Context Learning} Through {Multiple-Label
Words} in {Demonstration}.''} June 16, 2024.
\url{http://arxiv.org/abs/2406.10908}.

\end{CSLReferences}
\end{document}